\documentclass[10pt,twocolumn,letterpaper]{article}

\usepackage[algorithms]{wacv}      

\definecolor{wacvblue}{rgb}{0.21,0.49,0.74}
\usepackage{graphicx}
\usepackage{amsmath}
\usepackage{amssymb}
\usepackage{booktabs} 
\usepackage{multirow} 
\usepackage{array}    
\usepackage{siunitx}  
\usepackage{placeins} 
\usepackage{afterpage} 
\newcolumntype{P}[1]{>{\centering\arraybackslash}p{#1}}

\usepackage[pagebackref,breaklinks,colorlinks,allcolors=wacvblue]{hyperref}

\usepackage[capitalize]{cleveref}
\crefname{section}{Sec.}{Secs.}
\Crefname{section}{Section}{Sections}
\Crefname{table}{Table}{Tables}
\crefname{table}{Tab.}{Tabs.}

\def\wacvPaperID{228} 
\def\confName{WACV}
\def\confYear{2026}

\begin{document}

\title{FSMODNet: A Closer Look at Few-Shot Detection in Multispectral Data}

\author{
Manuel Nkegoum${}^{1,3}$, Minh-Tan Pham${}^1$,  Élisa Fromont${}^2$, Bruno Avignon${}^3$ and Sébastien Lefèvre${}^{1,4}$\\ \\
${}^1$Univ Bretagne Sud, IRISA, UMR 6074, Vannes, France\\
${}^2$Univ Rennes, IRISA, UMR 6074, Rennes, France\\
${}^3$ATERMES, Montigny-le-Bretonneux, France \\
${}^4$UiT The Arctic University of Norway, Tromsø, Norway
}
\maketitle

\begin{abstract}
Few-shot multispectral object detection (FSMOD) addresses the challenge of detecting objects across visible and thermal modalities with minimal annotated data. In this paper, we explore this complex task and introduce a framework named "FSMODNet" that leverages cross‑modality feature integration to improve detection performance even with limited labels. By effectively combining the unique strengths of visible and thermal imagery using deformable attention, the proposed method demonstrates robust adaptability in complex illumination and environmental conditions. Experimental results on two public datasets show effective object detection performance in challenging low-data regimes, outperforming several baselines we established from state-of-the-art models. All code, models, and experimental data splits can be found at \href{https://anonymous.4open.science/r/Test-B48D}{https://anonymous.4open.science/r/Test-B48D}.

\begin{figure}[ht!]
    \centering
    \includegraphics[width=0.44\linewidth]{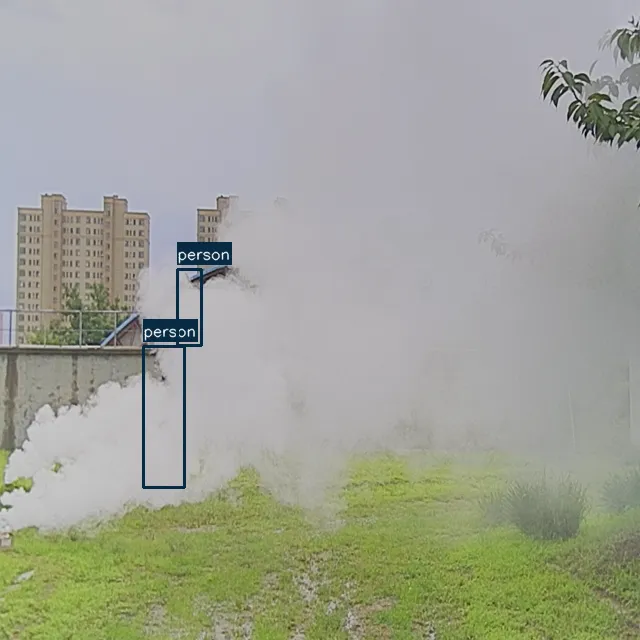}
    \includegraphics[width=0.44\linewidth]{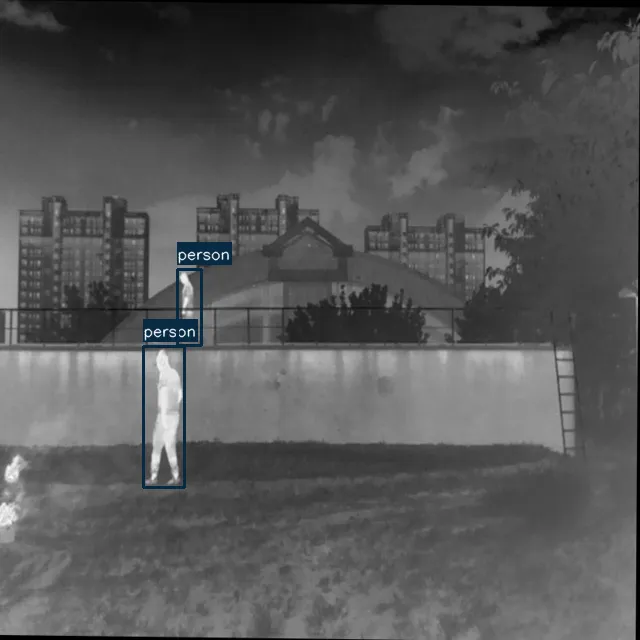}
    \includegraphics[width=0.44\linewidth]{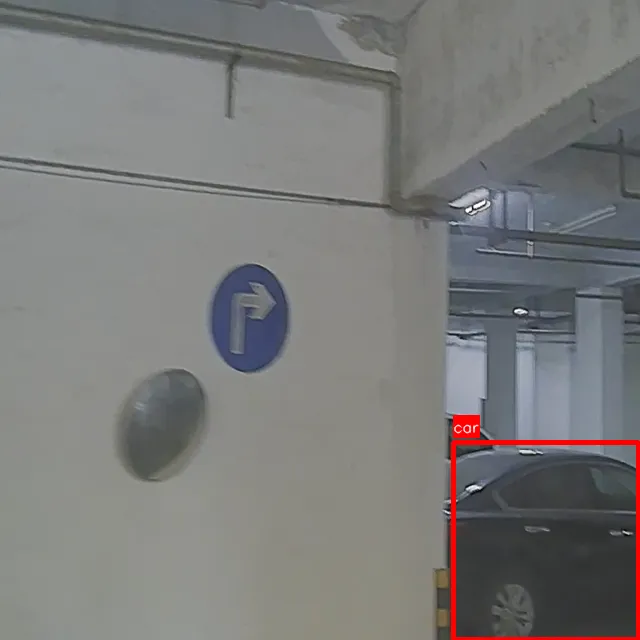}
    \includegraphics[width=0.44\linewidth]{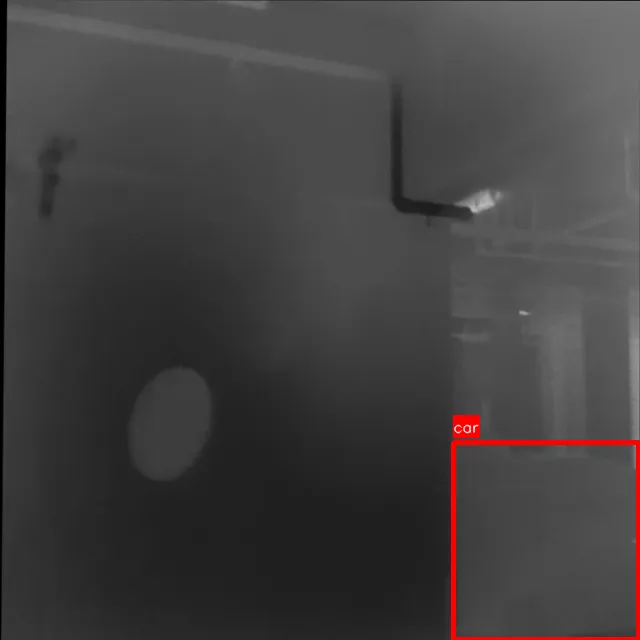}
    \caption{Two challenging multispectral scenes (top and bottom) with RGB (left) and IR (right) images from the M3FD dataset. Instances of the pedestrian and car classes are shown in top and bottom lines, respectively.}
    \label{fig:two_figures}
\end{figure}
\end{abstract}

\section{Introduction}
\label{sec:intro}
Object detection is a cornerstone of modern computer vision, powering applications from autonomous vehicles to security systems. While most traditional methods rely on RGB images, their performance often degrades in adverse conditions, such as low light, fog, or occlusion, as illustrated in \cref{fig:two_figures}. Multispectral object detection (MOD), which fuses visible and thermal modalities, offers a compelling alternative by leveraging both visual appearance and heat signatures. However, the success of MOD has so far hinged on large annotated datasets, which are costly and time-consuming to obtain.
Few-shot object detection (FSOD), which aims to detect novel object classes from only a few labeled examples, offers a promising avenue to alleviate this annotation burden. Yet, the intersection of FSOD and MOD, few-shot multispectral object detection (FSMOD), remains largely unexplored. To the best of our knowledge, only one recent study~\cite{huang2024cross} has attempted to address this setting  without releasing code or datasets nor a strong evaluation of the components of the proposed method.

In this work, we provide a first reproducible baseline for FSMOD closing key data and implementation gaps. Our evaluation is conducted on the FLIR~\cite{flir2019} and M³FD~\cite{liu2023m3fd} datasets. In particular, we propose a modular baseline architecture for FSMOD incorporating state-of-the-art fusion components and show its superior performance compared to~\cite{huang2024cross}.

In Section \ref{sec:related}, we cover related work ranging from simple object detection to multispectral few-shot object detection. In Section \ref{sec:method} we propose a new fusion architecture that, combined with a prototype aggregation architecture, can effectively tackle FSMOD. The performance of FSMODNet is evaluated in Section \ref{sec:expe}. We conclude in Section \ref{sec:conclusion}.


\section{Related Work}
FSMOD stands at the apex of a long line of research dedicated to object detection in outdoor environments. In what follows, we present the key methods that form the foundations of this dynamic and well-established domain, and which are essential to address the challenges of this emerging yet highly valuable task.

\label{sec:related}
\subsection{Object Detection}
Object detection has evolved from the two-stage RCNN series to real-time and transformer-based models. RCNN~\cite{girshick2014rcnn} proposed a two-stage pipeline with region proposals and CNN-based classification, later improved by Fast R-CNN~\cite{girshick2015fast} and Faster R-CNN~\cite{ren2015faster} through the introduction of a region proposal network. In parallel, YOLO~\cite{redmon2016yolo} and SSD~\cite{liu2016ssd} offered single-shot detectors prioritizing speed. More recently, DEtection TRansformer, DETR~\cite{carion2020end} reformulated detection as a set prediction task, removing anchors and Non-Maximum Suppression (NMS). DAB-DETR~\cite{liu2022dab} added dynamic anchor refinement, and Deformable DETR~\cite{zhu2020deformable} used deformable attention for faster convergence. RT-DETR~\cite{zhao2024detrs} further optimized this design for real-time scenarios.

\subsection{Multispectral Object Detection}
Multispectral detection combines visible and infrared imagery, requiring effective feature fusion. Zhang \etal~\cite{zhang2021guided} introduced auxiliary pedestrian masks to guide intra- and inter-modality learning. Transformer-based models~\cite{qingyun2021cross} integrate self-attention to capture local-global dependencies across modalities. Guo \etal~\cite{guo2025damsdet} address misalignment via deformable cross-attention, enhancing spectral feature aggregation. Zhang \etal~\cite{zhang2024when} propose a unified framework using a modality-adaptive transformer and query selection to filter redundant information. These methods highlight the role of adaptive fusion and attention in improving multispectral detection under challenging conditions.

\subsection{Few-shot Object Detection}
FSOD tackles detection with limited labeled data. Meta-DETR~\cite{zhang2022meta} applies meta-learning to generalize across classes without region proposals. FS-DETR~\cite{bulat2023fs} introduces visual prompts for novel-class inference. Wang \etal~\cite{wang2020tfa} show that fine-tuning only the final layer achieves strong performance. FSCE~\cite{Sun_2021_CVPR} enhances feature discrimination via contrastive proposal encoding. DE-ViT~\cite{pmlr-v270-zhang25i} improves localization by projecting transformer features into a subspace less prone to overfitting. Foundation-based FSOD~\cite{Han_2024_CVPR} leverages multimodal in-context learning, using language-guided examples to simplify and strengthen detection.

\begin{figure*}[t]
\centering
\includegraphics[width=\linewidth,keepaspectratio]{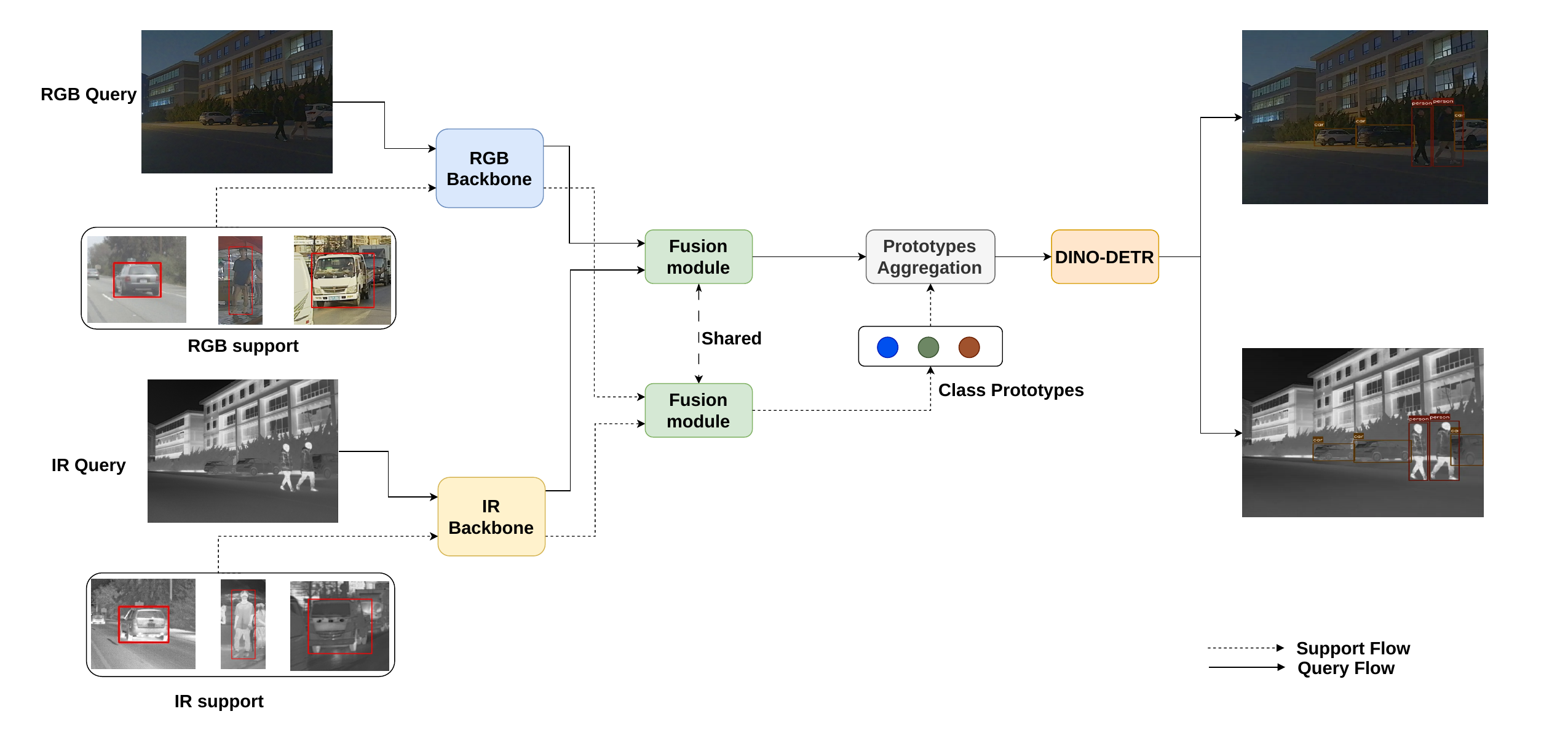}
\caption{The overall architecture of the proposed FSMODNet framework. Each modality (RGB and IR) is processed through dedicated encoders for both the support and query branches. Extracted features are then fused using our multi-stage fusion module that includes intra-spectrum enhancement via Neighborhood Attention and inter-spectrum interaction through Cross-Deformable Attention. Support prototypes refine the fused features via an attention mechanism before final detection using a DINO-based head.}
\label{fig:arch}
\end{figure*}

\subsection{Few-shot Multispectral Object Detection}
To the best of our knowledge, FSMOD was addressed in a single study so far. In their recent work, Huang et al.~\cite{huang2024cross} propose a Cross‑Modality Interaction (CMI) module, with spatial and channel attention, to fuse visible and thermal features, and a Semantic Prototype Metric (SPM) loss that leverages word embeddings for improved category separability. Despite these innovations, their approach has several shortcomings: (1) no monospectral baselines are reported, making it impossible to isolate the benefit of multispectral fusion under few‑shot regimes; (2) training and fine‑tuning are performed on different datasets, introducing a domain gap due to sensor, illumination, and distribution discrepancies; (3) evaluation is limited to 1‑, 2‑, 3‑, and 5‑shot settings, leaving performance at higher shot counts (e.g., 10‑ or 30‑shot) unexplored; and (4) neither code nor datasets are publicly released, hindering reproducibility and further comparative study.

\section{Methodology}
\label{sec:method}
Few-shot learning in general and FSOD in particular involves two distinct sets of classes: \( C_{\text{base}} \) and \( C_{\text{novel}} \), where \( C_{\text{base}} \cap C_{\text{novel}} = \emptyset \). Given a base dataset \( D_{\text{base}} \) with abundant samples and a novel dataset \( D_{\text{novel}} \) with only \( K \) annotated instances per class, the goal is to train a detector capable of identifying objects from both sets. FSOD aims to leverage a limited number of visual examples to detect novel classes while utilizing abundant training data from base classes. The objective is to achieve accurate detection of novel categories while maintaining strong performance on the base classes. The training is organized into episodic tasks that involve two fundamental components. The \textbf{support} which is a small set of annotated examples for each novel class that serves as the reference or prototype for the detection task, and the \textbf{query} image in which the detector is required to locate and classify objects, leveraging the information provided by the support set.

\subsection{Framework Overview}
The structure of our model is illustrated in \cref{fig:arch}. We detail here its different components, and especially our original fusion module.\\
\textbf{Feature extraction}
Specifically, we use one encoder per spectral modality that is shared between support and query sets to extract features at different semantics levels. Subsequently, the encoded features of each modality are passed through the fusion module for both query and support set. \\
 \textbf{Fusion module for multispectral feature integration}
Let $F_{IR}, F_{RGB} \in \mathbb{R}^{D \times H \times W}$ denote the feature maps (extracted from the IR and RGB modalities) where $D$ is the depth of the feature maps and $H \times W$ are the spatial dimensions. Our fusion module is composed of three sequential stages: intra-spectrum processing using Neighborhood Attention~\cite{hassani2023neighborhood} as implemented in the NATTEN package~\cite{NATTEN}, inter-spectrum processing via bidirectional cross-deformable attention, and a final fusion using a $1 \times 1$ convolution. To compute neighborhood attention, we first project the feature map (e.g., $F_{RGB}$) into queries, keys, and values using $1 \times 1$ convolutions:
\begin{equation}
    Q = W_Q * F_{RGB}, \quad K = W_K * F_{RGB}, \quad V = W_V * F_{RGB},
\end{equation}
where $Q, K, V \in \mathbb{R}^{D \times H \times W}$. For each spatial location $(i, j)$, we define, following~\cite{hassani2023neighborhood}, a local neighborhood $\mathcal{N}_{(i,j)}$, for example a $k \times k$ window centered at $(i, j)$. Attention is computed only within this neighborhood. The attention score between location $(i,j)$ and a neighbor $(u,v) \in \mathcal{N}_{(i,j)}$ is given by the scaled dot product:
\begin{equation}
\alpha_{(i,j),(u,v)} = \frac{Q_{i,j} \cdot K_{u,v}}{\sqrt{D}}.
\end{equation}
 These scores are normalized with a softmax over the neighborhood:
\begin{equation}
A_{i,j,u,v} = \frac{\exp(\alpha_{(i,j),(u,v)})}{\sum\limits_{(u',v') \in \mathcal{N}_{(i,j)}} \exp(\alpha_{(i,j),(u',v')})}.
\end{equation}
The output at location $(i,j)$ is computed as the weighted sum of the corresponding values:
\begin{equation}
F'_{RGB,i,j} = \sum_{(u,v) \in \mathcal{N}_{(i,j)}} A_{i,j,u,v} \cdot V_{u,v}.
\end{equation}

\begin{figure*}[t]
\centering
\includegraphics[width=\linewidth,keepaspectratio]{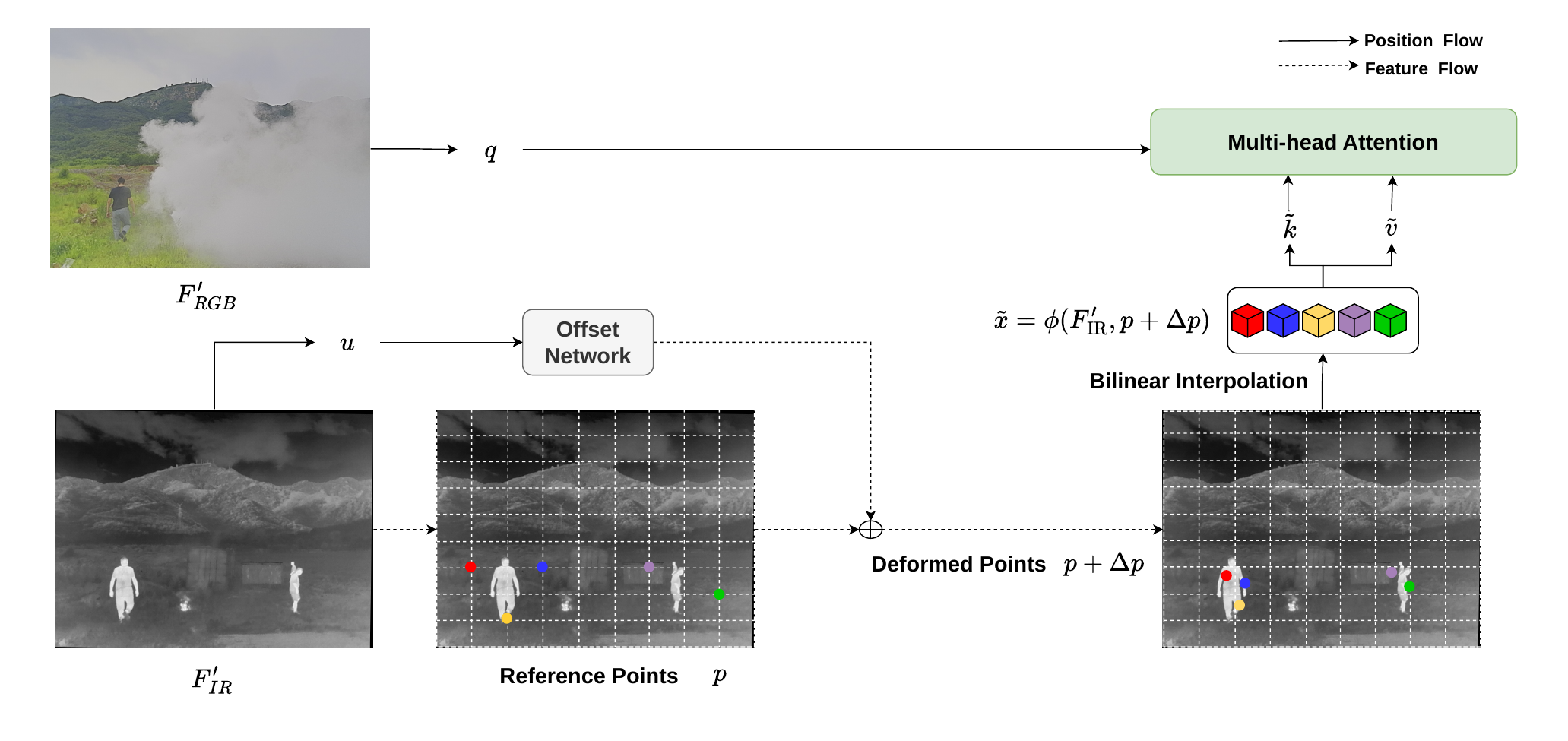}
\caption{An illustration of Cross-Deformable Attention mechanism. A group of reference points $p$ is placed uniformly on the feature map ($F'_{IR}$ for example), whose offsets $\Delta{p}$ are learned by the offset generation network. The features of important regions are sampled according to the deformed points with bilinear interpolation $\phi$. The deformed keys $\tilde{k}$ and values $\tilde{v}$ are projected by the sampled features and then participate in computing attention. We show only 5 reference points for a clear display. The offset network consists of a depth-wise convolution with stride $r$ to downsample the feature map and a $1 \times 1$ convolution following normalization and activation transforms the feature map to the offset value.}
\label{fig:dca}
\end{figure*}

The resulting output feature map $F_{RGB}'$ contains locally aggregated information with attention-based weighting, offering a more efficient and spatially-aware alternative to global self-attention.  
After intra‑spectrum processing, we obtain refined feature maps \(F'_{IR}\) and \(F'_{RGB}\). To fuse complementary cues and correct spatial misalignment between modalities, we propose a Cross‑Deformable Attention module (see \cref{fig:dca}). In this mechanism, queries are drawn from one modality while keys and values come from the other. For instance, when updating RGB features, the query is computed from \(F'_{RGB}\) and the keys/values are sampled from \(F'_{IR}\).
Inspired by the deformable attention module of~\cite{xia2022vision}, we begin by defining a uniform grid of reference points \( p \in \mathbb{R}^{2 \times H_G \times W_G} \) based on the IR feature map \( F'_{\text{IR}} \). Here, \( H_G = H / r \) and \( W_G = W / r \) denote the downsampled spatial dimensions with a factor \( r \). These reference points are normalized to the range \([-1, +1]\), mapping \((-1, -1)\) to the top-left corner and \((+1, +1)\) to the bottom-right corner of the feature map. For each reference point \(p\), we first extract its feature vector by sampling the IR feature map at \(p\) and projecting it via $u_p = W_uF'_{\text{IR}}$ then feed \(u_p\) into the lightweight network \(\theta_{\text{offset}}\) comprising a $k\times k$ depth‑wise convolution with stride $r$ (where $k > r$ to ensure a sufficient receptive field), followed by LayerNorm and GELU activation and a $1\times1$ convolution to predict the 2‑D offsets $\Delta p$. However and differently from \cite{xia2022vision}, in our multispectral case, when using the bilinear interpolation, we sample features from the important regions of IR feature map at these deformed locations as keys and values, followed by projection matrices:
\begin{align}
    \Delta p &= s \cdot \tanh(\theta_{\text{offset}}(u)), \tilde{x} = \phi(F'_{\text{IR}}, p + \Delta p) , \\
    \tilde{k} &= W_k \cdot \tilde{x},  \quad \tilde{v} = W_v \cdot \tilde{x},
\end{align}
where \( s \) is a predefined scaling factor to control the magnitude of the offsets and $\phi$ is the bilinear interpolation function. The query features are computed from the RGB feature map via a learnable projection matrix and the attention weights are computed:
\begin{align}
    q &= W_q \cdot F'_{\text{RGB}}, \\
    A &= \operatorname{Softmax}(\exp(q \cdot k/ \sqrt{D}).
\end{align}
The RGB feature map is then updated using these attention weights:
\begin{equation}
F_{\text{RGB}}''= F_{\text{RGB}} +  \text{ConvFFN}(F'_{\text{RGB}} + A \cdot v),
\end{equation}
where ConvFFN is a convolutional forward network consisting of two successive $1 \times 1$ convolutional layers with nonlinear activation (e.g., ReLU) in between, designed to refine spatial feature maps while preserving their spatial structure. An analogous formulation is applied for updating the infrared features using queries derived from $F'_{IR}$ and keys/values computed from $F'_{RGB}$. The final fusion step concatenates $F''_{IR}$ and $F_{RGB}''$ along the channel dimension (resulting in a feature map of size $\mathbb{R}^{2D \times H \times W}$) and processes them using a pointwise convolution:
\begin{align}
    F_q = \text{Conv}_{1 \times 1}(\text{cat}(F''_{IR},F''_{RGB})).
\end{align}%

\textbf{Prototypes aggregation}
We use Meta‑DETR~\cite{zhang2022meta} as our baseline network to build our meta‑learning scheme and recall here its main principles. The support class prototypes are first extracted by applying RoIAlign~\cite{he2017mask} to each ground‑truth support box, followed by average pooling to obtain one prototype per class. These prototypes form a set \(S \in \mathbb{R}^{C \times D}\), where \(C\) is the number of support classes. To address the class‑agnostic nature of meta‑learning, Meta‑DETR introduces a set of fixed task encodings \(T \in \mathbb{R}^{C \times D}\) for each class. Given the flattened query features \(F_q \in \mathbb{R}^{HW \times D}\), the class‑specific support prototypes \(S\), and predefined task‑level encodings \(T\), we first project them into a shared feature space via a learnable projection matrix \(W\). Subsequently, a single‑head self‑attention module is applied, where the support features serve as the key and value, and the query features act as the query. We recall here the main equations from \cite{zhang2022meta} not represented in Fig.~\ref{fig:arch}, starting with the attention
\begin{equation}
A = \text{Softmax}\!\left( \frac{(F_q W)(S W)^{\top}}{\sqrt{d}} \right).
\end{equation}
This attention is then used to generate two representations:
\begin{align}
Q_F &= A \odot \sigma(S), \\
Q_E &= A\,T,
\end{align}
where \(\odot\) denotes element‑wise multiplication and \(\sigma\) is the sigmoid activation function. \(Q_F\) selectively filters the query features using class‑specific information, while \(Q_E\) injects task‑level encoding into the query. The combined result is passed through a feed‑forward network (FFN) to produce the aggregated features:
\begin{equation}
F_{\text{CAM}} = \text{FFN}(Q_F + Q_E).
\end{equation}
This parallel attention branch enables the transformation of class‑specific support representations into a class‑agnostic prototype embedding.\\
\textbf{Detection} 
Afterwards, the final features are passed to DINO‑DETR~\cite{dino} to produce the final detection outputs. Additionally, following Meta R‑CNN~\cite{yan2019meta}, we employ a cosine similarity–based cross‑entropy loss~\cite{chen2019closer} to supervise prototype classification, encouraging strong alignment between the task prototype and the query features.  
\subsection{Training and Inference}
The proposed training framework consists of two stages: a base meta-learning stage and a few-shot fine-tuning stage. 
\textbf{Base Meta-Learning}  
We first train on the base dataset \(D_{\text{base}}\) enabling the model to learn dynamic class representations that generalize to novel categories. \\
\textbf{Few-Shot Fine-Tuning}  
In the second stage, the model is fine-tuned to recognize novel categories \( C_{\text{novel}} \) alongside the base categories \( C_{\text{base}} \). The novel dataset \( D_{\text{novel}} \) contains only \( K \) annotated examples per novel class, simulating the \( K \)-shot setting. To prevent catastrophic forgetting and maintain performance on the base classes, samples from \( D_{\text{base}} \) are also included during fine-tuning and follow the same episodic setting as the base learning scheme.\\
\textbf{Inference}
After the training phase, class prototypes can be precalculated once and for all by taking the average prototype of each class to reduce their variance. At test time, we use those prototypes for detection in each RGB/IR image pair.

\begin{table*}[t]
\small
\centering
\begin{tabular}{@{}lcc*{9}{S[table-format=2.2]}@{}} 
\toprule
\multirow{2}{*}{Method} & \multirow{2}{*}{Modality} & \multicolumn{3}{c}{Split 1} & \multicolumn{3}{c}{Split 2} & \multicolumn{3}{c}{Split 3} \\
\cmidrule(lr){3-5} \cmidrule(lr){6-8} \cmidrule(l){9-11} 
 &  & {5-shot} & {10-shot} & {30-shot} & {5-shot} & {10-shot} & {30-shot} & {5-shot} & {10-shot} & {30-shot} \\
\midrule
Meta-DETR~\cite{zhang2022meta}         & RGB & 3.77  & 6.26  & 8.44  & 4.58  & 4.50  & 6.67  & 1.16  & 3.98  & 7.38  \\
Meta-DETR~\cite{zhang2022meta}        & IR  & 3.28  & 5.73  & 10.07 & 1.67  & 3.96  & 3.94  & 1.48  & 4.35  & 11.27 \\
DAMSDet+TFA~\cite{guo2025damsdet,wang2020tfa}        & RGB+IR & 19.80 & 26.66 & 38.61 & 18.60 & 25.01 & 35.63 & 17.03 & 20.31 & 27.29 \\
CAFF-DINO+TFA~\cite{helvig2024caff,wang2020tfa}      & RGB+IR & 20.71 & 27.37 & 41.01 & 18.72 & 23.69 & 34.17 & 14.32 & 17.90 & 24.29 \\
\midrule
FSMODNet               & RGB    & 16.78 & 24.04 & 33.34 & 3.56  & 6.33  & 18.21 & 9.49  & 20.06 & 23.63 \\
FSMODNet               & IR     & \underline{26.60} &\underline{33.92} & \underline{46.25} & \underline{26.32} & \textbf{30.31} & \underline{38.05} & \underline{21.49} & \underline{29.74} & \underline{32.41} \\
FSMODNet               & RGB+IR & \textbf{30.67} & \textbf{36.56} & \textbf{49.70} & \textbf{27.05} & \underline{29.76} & \textbf{42.47} & \textbf{32.04} & \textbf{37.20} & \textbf{46.18} \\
\bottomrule
\end{tabular}
\caption{Few-shot object detection performance on novel classes (nAP50) of the FLIR dataset. We use ResNet-50 for feature extraction.}
\label{tab:results 1}
\end{table*}

\begin{table*}[t]
\small
\centering
\begin{tabular}{@{}lcc*{9}{S[table-format=2.2]}@{}} 
\toprule
\multirow{2}{*}{Method} & \multirow{2}{*}{Modality} & \multicolumn{3}{c}{Split 1} & \multicolumn{3}{c}{Split 2} & \multicolumn{3}{c}{Split 3} \\
\cmidrule(lr){3-5} \cmidrule(lr){6-8} \cmidrule(l){9-11} 
 &   & {5-shot} & {10-shot} & {30-shot} & {5-shot} & {10-shot} & {30-shot} & {5-shot} & {10-shot} & {30-shot} \\
\midrule
Meta-DETR~\cite{zhang2022meta}         & RGB    & 7.68  & 14.41 & 20.26 & 8.17  & 16.35 & 19.62 & 7.31  & 12.50 & \multicolumn{1}{c}{17.52} \\
Meta-DETR~\cite{zhang2022meta}       & IR     & 7.49  & 13.84 & \multicolumn{1}{c}{17.02} & 8.48  & 15.38 & \multicolumn{1}{c}{17.87} & 7.22  & 14.02 & \multicolumn{1}{c}{15.96} \\
DAMSDet+TFA~\cite{guo2025damsdet,wang2020tfa}    & RGB+IR & \underline{27.93} & 36.75 & \underline{46.62} & \underline{30.28} & 41.14 & \underline{49.95} & 24.78 & 37.75 & 45.92 \\
CAFF-DINO+TFA~\cite{helvig2024caff,wang2020tfa}    & RGB+IR & 18.31 & 27.08 & 39.58 & 19.55 & 33.04 & 42.95 & 19.78 & 32.72 & 41.57 \\
\midrule
FSMODNet               & RGB    & 27.62 & \underline{38.73} & 45.21 & 28.55 & \underline{42.55} & 47.58 & \textbf{27.75} & \textbf{44.11} & \underline{47.85} \\
FSMODNet               & IR     & 20.70 & 28.10 & 34.31 & 20.72 & 32.11 & 38.21 & 18.97 & 29.72 & 33.59 \\
FSMODNet               & RGB+IR & \textbf{29.43} & \textbf{40.11} & \textbf{47.32} & \textbf{37.02} & \textbf{47.22} & \textbf{52.18} & \underline{25.14} & \underline{42.71} & \textbf{48.77} \\
\bottomrule
\end{tabular}
\caption{Few-shot object detection performance on novel classes (nAP50) of the M³FD  dataset. We use ResNet-50 for feature extraction.}
\label{tab:results 2}
\end{table*}

\section{Experimental Results}
\label{sec:expe}
\subsection{Experimental Setting}
We use ResNet50~\cite{he2016deep} as the backbone for both the infrared and visible branches. The encoder and the decoder contain six layers both. The model is trained on 4 NVIDIA A100 GPUs using the AdamW~\cite{loshchilov2017decoupled} optimizer, with an initial learning rate of \(1 \times 10^{-4}\) and a weight decay of \(1 \times 10^{-4}\). The batch size is set to 16. During the base training stage, the model is trained for 50 epochs. During the fine-tuning stage, the learning rate is reduced to \(5 \times 10^{-5}\) and the model is trained for 100 epochs with a batch size of 2.
\subsubsection{Datasets}
In this study, we evaluate our FSMOD framework on two challenging datasets. 
\textbf{FLIR}~\cite{flir2019}: Aligned RGB-thermal pairs (4129 train / 1013 test) with three classes (people, car, bicycle). We evaluate three base/novel splits: (bicycle, car/people), (people, car/bicycle), (people, bicycle/car).\\
\textbf{M\textsuperscript{3}FD}~\cite{liu2023m3fd}: 4200 co‑registered RGB-thermal pairs across Day, Night, Overcast, and Challenge conditions with six classes (people, car, truck, bus, motorcycle, lamp). Using the dataset from~\cite{guo2025damsdet}, we form three base/novel splits by selecting the least frequent classes as novel: (people, car, lamp/others), (people, car, lamp, truck/others), (people, car, lamp, truck, bus/motorcycle).

\subsubsection{Few‑Shot Setting}
To comprehensively evaluate the effectiveness of our approach, we generate 10 distinct support datasets during the fine-tuning phase. The number of shots is set to 5, 10 and 30. The mean average precision of novel categories at IoU threshold 0.5 (nAP50) is used as the evaluation metric.

\subsubsection{Baselines}
Due to the lack of available FSMOD baselines for direct comparison, we construct our baselines by combining state-of-the-art MOD frameworks, \textbf{DAMSDet}~\cite{guo2025damsdet} and \textbf{CAFF-DINO}~\cite{helvig2024caff}, with the few-shot detection method \textbf{TFA}~\cite{wang2020tfa}. Let us note that \cite{huang2024cross} also consider TFA but with older MOD frameworks.
Specifically, during the base training phase, each model is first trained on the base classes. The backbone is then frozen, and the model is fine‑tuned on both base and novel classes. For CAFF‑DINO, we use its lite variant (CAFF*) replacing the original Swin‑Large backbone with ResNet‑50 to reduce computational cost and overfitting risks (that we observed in our preliminarily experiments). Additionally, we include Meta‑DETR in our evaluations, applying it independently to each monospectral modality (RGB and IR).

\subsection{Results}  
\begin{figure}
\centering
\includegraphics[width=\linewidth,keepaspectratio]{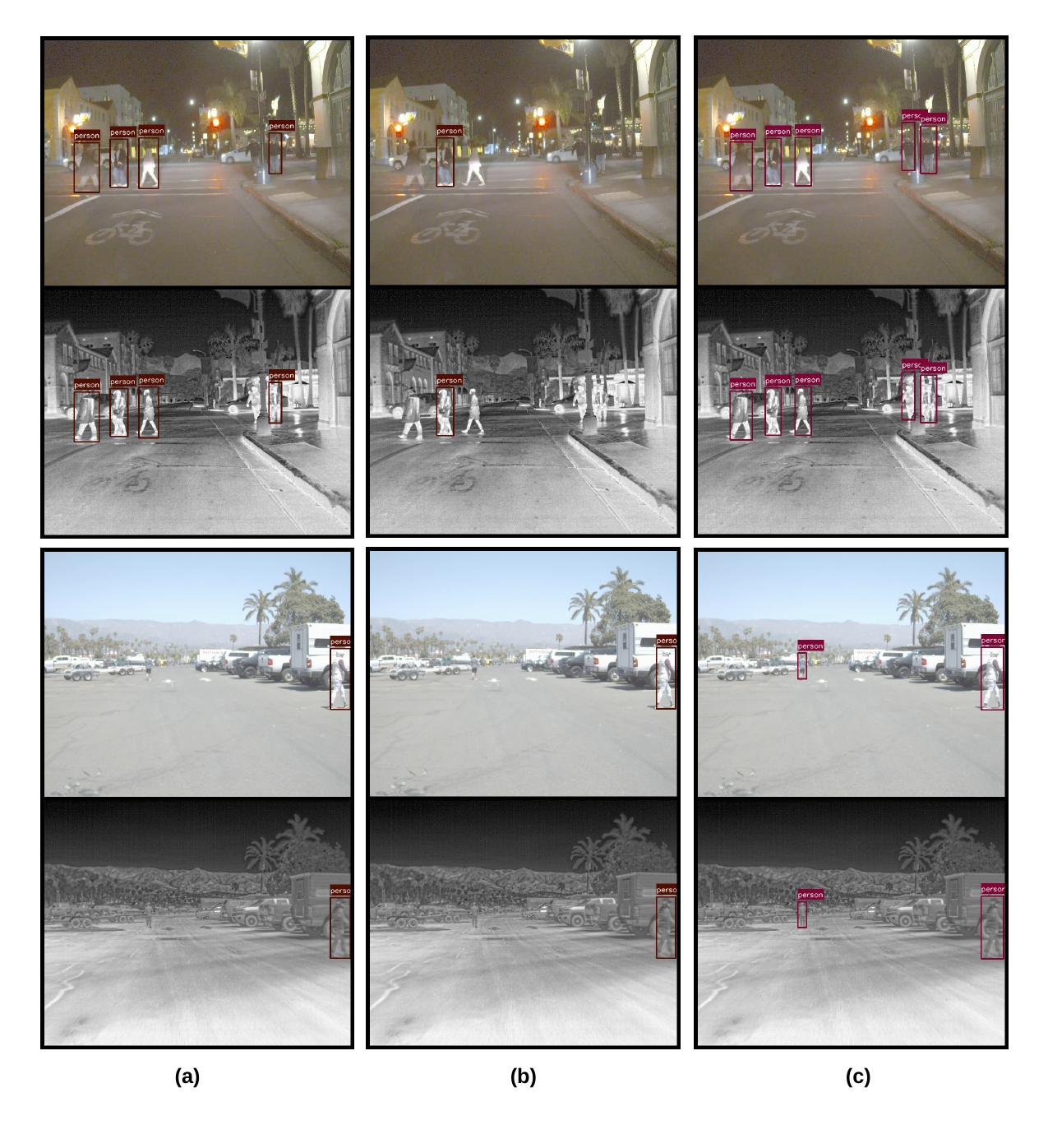}
\caption{Illustration of detection results on the FLIR dataset (Split 1), where the \textit{people} class is treated as novel. Models were fine-tuned with 30 shots. For each method, RGB (top) and IR (bottom) images are shown. (a) DAMSDet+TFA, (b) CAFF-DINO+TFA, (c) Our FSMODNet. Only detections of the novel class (\textit{people}) are visualized for clarity.}
\label{fig:vizu}
\end{figure}

\label{sec:results}

The results presented in \cref{tab:results 1,tab:results 2} highlight the strong performance of our proposed method across different dataset splits and few-shot scenarios. Below, we analyze the key findings and compare them with existing approaches.

\subsubsection{Impact of Multimodal Fusion}

Our method integrates RGB and infrared (IR) modalities to leverage complementary information for improved detection. While multimodal fusion generally yields superior performance across most splits and shot settings, the results in \cref{tab:results 1,tab:results 2} also reveal that monomodal models occasionally surpass the fused model in specific low-shot configurations. This phenomenon underscores the inherent difficulty of effectively optimizing multimodal fusion when training data is scarce. The fusion mechanism must learn to balance and integrate heterogeneous features from different modalities, which can be challenging in low-shot regimes where limited samples hinder robust joint representation learning especially for datasets where one modality is sufficient to perform detection as it is the case with the IR modality for FLIR.
Nonetheless, as the number of shots increases, the multimodal model consistently outperforms monomodal counterparts, demonstrating that the complementary nature of RGB and IR data can be effectively exploited given sufficient examples. For instance, on the FLIR dataset, the fused RGB+IR model attains up to 49.70 nAP50 at 30-shot in Split 1, surpassing both IR-only (46.25) and RGB-only (33.34) variants. 

\subsubsection{Few-Shot Learning Comparison}

Our method employs a meta-learning framework based on prototype representations, which enables more efficient adaptation to novel classes with limited samples. As shown in \cref{tab:results 1,tab:results 2}, our prototype-based meta-learning consistently outperforms TFA-based methods across all splits and shot configurations. For instance, on the FLIR dataset, our approach improves over DAMSDet+TFA by up to 11.09 nAP50 at 5-shot in Split 1 (30.67 vs 19.80) and sustains this advantage at higher shot counts. Similarly, on the M³FD dataset, our method achieves gains ranging from 1.5 to 6.74 nAP50 over DAMSDet+TFA in 5-shot scenarios, with the largest increase in Split 2 (37.02 vs 30.28). This performance gap highlights the superiority of prototype-based meta-learning in rapidly generalizing to new classes from few examples, compared to the feature adaptation strategy used in TFA. Furthermore, our approach exhibits steeper learning curves, demonstrating more effective utilization of additional shots to enhance detection accuracy.

\subsubsection{Ablation Studies on FLIR}

To better understand the contributions of different components in our method, we perform ablation studies on Split 1 of the FLIR dataset. We focus on two aspects: choice of backbone encoder and choice of fusion module. All experiments use our prototype-based meta‑learning framework with the same training protocol; only the backbones or fusion layers differ.

\textbf{Backbone Choice}
We adopt ResNet-50 as our default encoder and compare it against ConvNeXt‑S, Swin‑T, and DinoV2-S under 5‑, 10‑, and 30‑shot settings (Tab.~\ref{tab:backbone}). As DinoV2 is not inherently multiscale we built a multiscale variant using bilinear interpolation and convolutions. Transformer-based backbones like Swin‑T and DinoV2-S use dense self-attention to capture spatial context, but when paired with our fusion block which also contains self-attention, they can incur feature redundancy that can hinder prototype discrimination under few-shot constraints. ResNet50’s convolutional features, by contrast, complement the fusion block’s intra-modality attention without overlap, leading to more stable and informative prototypes from limited samples.
\begin{table}[h]
\small
\centering
\begin{tabular}{l
                S[table-format=2.2]
                S[table-format=2.2]
                S[table-format=2.2]}
\toprule
Backbone & {5-shot} & {10-shot} & {30-shot} \\
\midrule
ConvNeXt-S~\cite{liu2022convnet} & 22.34 & 30.59 & 41.83 \\
Swin-T~\cite{liu2021swin}        & 23.66 & 33.42 & 44.50 \\
DINOV2-S~\cite{oquab2024dinov}   & 25.30 & 33.21 & 44.67 \\
ResNet-50                        & \textbf{30.67} & \textbf{36.56} & \textbf{49.07} \\
\bottomrule
  \end{tabular}
  \caption{Backbone ablation study on Novel Split 1 of FLIR.}
  \label{tab:backbone}
\end{table}

\textbf{Fusion Approach}
We also evaluate three simpler fusion strategies against our proposed fusion block with Cross‑Deformable Attention (CDA) module. “Concat” concatenates RGB and IR feature maps followed by a $1\times1$ convolution, “Add” simply sums the two feature maps and "CMI" is the one proposed by~\cite{huang2024cross}. Results are presented in \cref{tab:fusion}. While simple concatenation approaches our performance at 5‑shot, it lags by over 4 points at 30‑shot. The additive fusion is less effective in low‑shot settings (5‑shot nAP50 of 22.82). In contrast, our CDA‑based fusion consistently outperforms both by effectively aligning and integrating cross‑spectrum features, confirming its superiority for robust multimodal few‑shot detection.

\begin{table}[h]
\small
\centering
\begin{tabular}{l
                S[table-format=2.2]
                S[table-format=2.2]
                S[table-format=2.2]}
\toprule
Fusion Method & {5-shot} & {10-shot} & {30-shot} \\
\midrule
Concat        & 29.62   & 33.38   & 45.40   \\
Add           & 22.82   & 32.67   & 44.95   \\
CMI~\cite{huang2024cross}  &  \textbf{31.52}  &    34.61     &   46.04 \\ 
CDA           & \textbf{30.67} & \textbf{36.56} & \textbf{49.07} \\
\bottomrule
  \end{tabular}
  \caption{Fusion approach comparison on Split 1 of FLIR (people as novel class). CDA: Cross‑Deformable Attention, CMI: Cross-Modality Interaction.}
  \label{tab:fusion}
\end{table}

\subsubsection{Runtime analysis}
In \cref{tab:arch_cost}, we compare model size, FLOPs, and FPS, observing that our framework incurs higher latency due to its episodic conditioning: each forward pass can detect only \(T\) classes, so detecting more classes requires multiple passes of the DETR head. This repeated decoding adds overhead compared to baselines that process all classes at once. However, in practical scenarios where only a few target classes are needed, restricting the support set to those classes allows single‑pass inference, significantly reducing latency and making the approach viable for real‑time use.

\begin{table}[h]
\small
\centering
\begin{tabular}{l
                S[table-format=3.0] 
                S[table-format=2.0] 
                S[table-format=2.2]}
\toprule
Method & {GFLOPs} & {Params (M)} & {FPS} \\
\midrule
DAMSDet+TFA~\cite{guo2025damsdet,wang2020tfa}   & 136 & 78 & 13.98 \\
CAFF-DINO+TFA~\cite{helvig2024caff,wang2020tfa} & 166 & 79 & 9.43  \\
FSMODNet                                         & 362 & 86 & 5.94  \\
\bottomrule
\end{tabular}
\caption{Architectural Cost Analysis. We use 640×640 resolution as input with FP32-precision on NVIDIA GeForce RTX 2080 Ti without post-processing. We use the FLIR dataset for experiments.}
\label{tab:arch_cost}
\end{table}
\subsubsection{Qualitative Results}
\cref{fig:vizu} illustrates qualitative comparison on Split 1 of the FLIR dataset under the 30-shot fine-tuning protocol, treating “people” as the novel class. For each scene, the top row shows the RGB input and the bottom row the corresponding IR input. Columns (a), (b), and (c) display detections by DAMSDet, CAFF-DINO, and our FSMODNet, respectively. All boxes are thresholded at IoU = 0.5 and only “people” detections are overlaid. Despite the extreme scarcity of labeled examples, our method (c) consistently identifies pedestrians of varying sizes and distances. In the first scene, FSMODNet correctly localizes two small figures at the image edges where both baselines miss or mislocalize one of them. In the second scene, while DAMSDet (a) produces several false alarms and CAFF-DINO (b) omits a distant walker, FSMODNet detects all true positives with minimal spurious boxes. This demonstrates that cross-modality fusion and prototype refinement enable our model to generalize robustly to challenging novel-class instances.

\section{Conclusion}
\label{sec:conclusion}
We present a reproducible framework for FSMOD that unifies RGB and infrared modalities within a principled episodic evaluation protocol. At the core of our approach is a novel multimodal fusion module, which leverages Neighborhood Attention for intra-spectrum feature refinement and bidirectional Cross-Deformable Attention for inter-spectrum alignment, all integrated into a meta-learning paradigm. Extensive experiments on the FLIR and M³FD datasets show that our fusion model consistently outperforms monomodal baselines, achieving up to 49.70 nAP50 at 30-shot on FLIR Split 1 (compared to 33.34 for RGB-only), and delivering gains of up to 11.09 nAP50 over TFA-based methods at 5-shot. Looking ahead, we plan to extend our fusion architecture to additional modalities (e.g., text), design adaptive task encoding strategies to reduce inference costs, and explore unsupervised pretraining methods tailored to multispectral few-shot learning.

{\small
\bibliographystyle{ieee_fullname}
\bibliography{ref.bib}
}

\end{document}